\documentclass{article} 
\usepackage{nips12submit_e,times}
\usepackage{graphicx}
\usepackage{subfigure}
\usepackage{amsmath}
\usepackage{amssymb}
\usepackage{hyperref}

\title{On Learning Where To Look}
\author{Marc'Aurelio Ranzato \\
Google Inc.\thanks{The author is currently with Facebook. This work was done while at Google.} \\
Mountain View CA, U.S.A.
}

\nipsfinalcopy 

\def\y{{\mathbf y}}
\def\z{{\mathbf z}}
\def\h{{\mathbf h}}
\def\b{{\mathbf b}}
\def\l{{\mathbf l}}

\def\L{{\cal L}}
\def\o{{\mathbf o}}

\begin{document}
\maketitle

\begin{abstract}
Current automatic vision systems face two major challenges:
scalability and extreme variability of appearance.
First, the computational time required to process
an image typically scales linearly with the number of pixels in the image,
therefore limiting the resolution of input images to thumbnail size. Second,
variability in appearance and pose of the objects constitute a major
hurdle for robust recognition and detection. In this work, we propose
a model that makes baby steps towards addressing these challenges. We
describe a learning based method that recognizes objects through a
series of glimpses. This system performs an amount of computation
that scales with the complexity of the input rather than its number of
pixels. Moreover, the proposed method is potentially more robust to
changes in appearance since its parameters are learned in a data driven manner.
Preliminary experiments on a handwritten dataset of digits demonstrate
the computational advantages of this approach.
\end{abstract}

\section{Introduction}
There are about 40 types of vision systems in nature~\cite{HumanAnimalVision}. Each such vision system
is highly optimized for certain physical constraints, for the
environment where the animal lives and for the task that it
has to accomplish in order to survive. Machine vision has made many
strides in the past decades with successful applications in several
domains. However when compared to human vision, artificial visual systems are
rather primitive in terms of their efficiency and ability to work in
unrestricted domains. We conjecture that if we aim at building an
artificial system that is good at tasks for which humans are good,
then it may be helpful to take inspiration from some of the principles underlying
the human visual system.

One striking difference between machine vision and human vision is how
we process the visual world. The human visual system is foveated,
meaning that its acuity (related to the density of cones in the retina) is much
larger around the fovea than other areas. Moreover, humans analyze the
visual world through a sequence of saccades. By integrating
information across the whole sequence of glimpses, humans form a mental
description of the visual world surrounding them.

Machine vision is currently more similar to the visual system of a bee
instead. Eyes do not move, but process every part of the image in the
same way. Local features are extracted at all locations and they are
aggregated across different spatial regions in order to yield a
representation of the whole image.
Such a processing strategy has two major limitations. First, it is not
scalable because the computational cost scales linearly with the number of
pixels in the image. Since modern cameras take photographs that
 easily have more than 10M
pixels, it would take several seconds if not minutes to analyze a
single image. Second, the representation produced by such a system is
{\em equivariant} as opposed to {\em invariant}, meaning that spatial
translation and changes of scale are going to be reflected in the low
level features that the system extracts. However, recognition tasks require
the representation to be robust to such changes.

Current state-of-the-art computer vision systems~\cite{kriz12} have addressed these
issues in several ways. First, images are often down-sampled to
thumbnail size (typical resolution is 200 by 200 pixels, for
instance). While this alleviates the scalability problem of
searching over too many locations and scales~\cite{Rowley}, it seems a rather unsatisfactory solution because it limits recognition to
objects that are at a rather coarse scale of the image.
Second, invariance is achieved by computing histograms of local
features over large regions at different locations and
scales~\cite{lazebnik-cvpr-06}.
While this may achieve the desired outcome, it seems wasteful if
not error prone to process regions of the image that do not
contain the object of interest.

Today, the most computationally efficient and best performing methods employ
the above mentioned strategies at their core.
The best image recognition system, namely the Convolutional Neural
Network~\cite{lecun-98,kriz12}, is typically trained on small image patches and then
``unrolled'' over larger images at test time~\cite{Osadchy}. The key computation is
the application of a sequence of stages, each composed by a non-linear
filter bank across all spatial locations and scales, followed by some spatial pooling transform.
Similarly, the best detection method, namely the Deformable Parts
Model~\cite{DPM}, applies a filter bank at all locations and scales,
followed by a cascade of similar operations at the most promising
locations in order to cut the computational cost~\cite{viola-01}.

In the past years, there have been several attempts at building a
radically different paradigm of visual processing based on a foveated
vision. However, most work in this area has been restricted to specific
domains~\cite{Elder}, to rather artificial tasks~\cite{Renninger} and it has not
been competitive on standard benchmark datasets~\cite{Larochelle-fovea}.

In this work, we propose a method to learn where to
look sequentially in images. The key idea is to first aggressively
down-sample the image to cheaply detect candidate locations where to look at
higher resolution. At test time, the information about low level
features as well as the class distribution produced by this system are
used to predict the location where to look next at high resolution.
Given this location, a patch is cropped at the predicted location and
it is fed to another system that predicts a new distribution over the
classes. This process repeats for as many glimpses as necessary to
reduce the uncertainty about the category of the object in the
image. Information across the glimpses is integrated by taking a
geometric mean of the intermediate predictions.
At training time, the location of the glimpse is treated as a latent
variable. Given a prediction of the location, we perturb this estimate
and perform a local discrete search to find a better location.

Our empirical validation shows that this approach is able to achieve
very competitive results on variants of the MNIST dataset~\cite{MNIST} of
handwritten digits. These experiments show that we can drastically
cut the computational cost of classification while maintaining
or even improving accuracy. The system does not waste computation
in parts of the image that are unlikely to contain informative
features and it does not waste computation on inputs that are simple
enough to be classified confidently with few glimpses.

\section{Algorithm}

In this section, we will discuss how the system works at test and
training time. In the next section instead, we will discuss how this
relates to other methods.

\subsection{Inference}
Given an input image $X \in R^{D \times D}$, we want to predict
its class label $y \in [0,1, \dots , C]$.
Notice that $D$ may be large and therefore densely processing the whole
image may be too expensive.
In this work, we are going to consider multi-layer fully connected
neural networks as classification methods. However, notice that
this approach can be used with any other classifier as well.

The first step is to aggressively down-sample the image to a much
lower resolution: $\bar{X} \in R^{d \times d}$ with $d \ll D$.
Then, $\bar{X}$ is fed to a fully connected neural network. Such
network can have as many hidden layers as needed. For simplicity,
here we assume that the network has only one hidden layer.

The transformations produced by this network are:
\begin{eqnarray}
\begin{array}{lll}
\h^{L} & = & \max(0, W_1^L \bar{X} + \b_1^L) \\
\o^{L} & = & W_2^L \h^L + \b_2^L
\end{array}
\end{eqnarray}
where $W_1^L$ and $W_2^L$ are weight matrices and $\b_1^L$ and
$\b_2^L$ are biases. These are also the parameters that are subject
to learning at training time. The output of the network, $\o^{L}$,
is a $C$ dimensional vector. By passing this vector through a
softmax non-linearity we produce a distribution over the classes:
\begin{equation}
\y^L_k = \mbox{softmax}(\o^{L}_k) = \frac{\exp(\o^{L}_k)}{ \sum_j
\exp(\o^{L}_j) } \label{eq:prob-lowreznet}
\end{equation}
since each component of this vector is positive and components
sum to one.
We call this network the ``low-resolution network'' (N0), since it
processes a down-sampled version of the original image.
Next, the low level features $\h^{L}$ and the output of the
network $\o^L$ are fed to another neural network that predicts the
(normalized) (x,y) coordinates of where to look next at high
resolution. For simplicity but without loss of generality, here we
consider a simple sigmoid layer:
\begin{equation}
\l^{p_1} = \sigma(W^{p_1}_1 \h^{L} + W^{p_1}_2 \o^{L} + \b^{p_1})
\end{equation}
where $\sigma(x) = 1 / (1 + \exp(-x))$. Again, $W^{p_1}_1$, $W^{p_1}_2$ and $\b^{p_1}$ are parameters which are subject to learning at training time.
The predicted location, $\l^{p_1}$
is a vector with two components between 0 and 1, specifying the
(normalized) coordinates of the location where to look next at
high resolution.
 More generally, we could also predict the width
and the height of the patch as well as other properties.

At the next step, we crop a patch at the
location specified by $\l^{p_1}$ from the high resolution image
$X$, we denote this by $X(\l^{p_1}) \in R^{w \times w}$.
 In this work, the patch is square and its diameter is fixed in advance to
$w$.

\begin{figure}[t]
\begin{center}
\includegraphics[width=.5\linewidth]{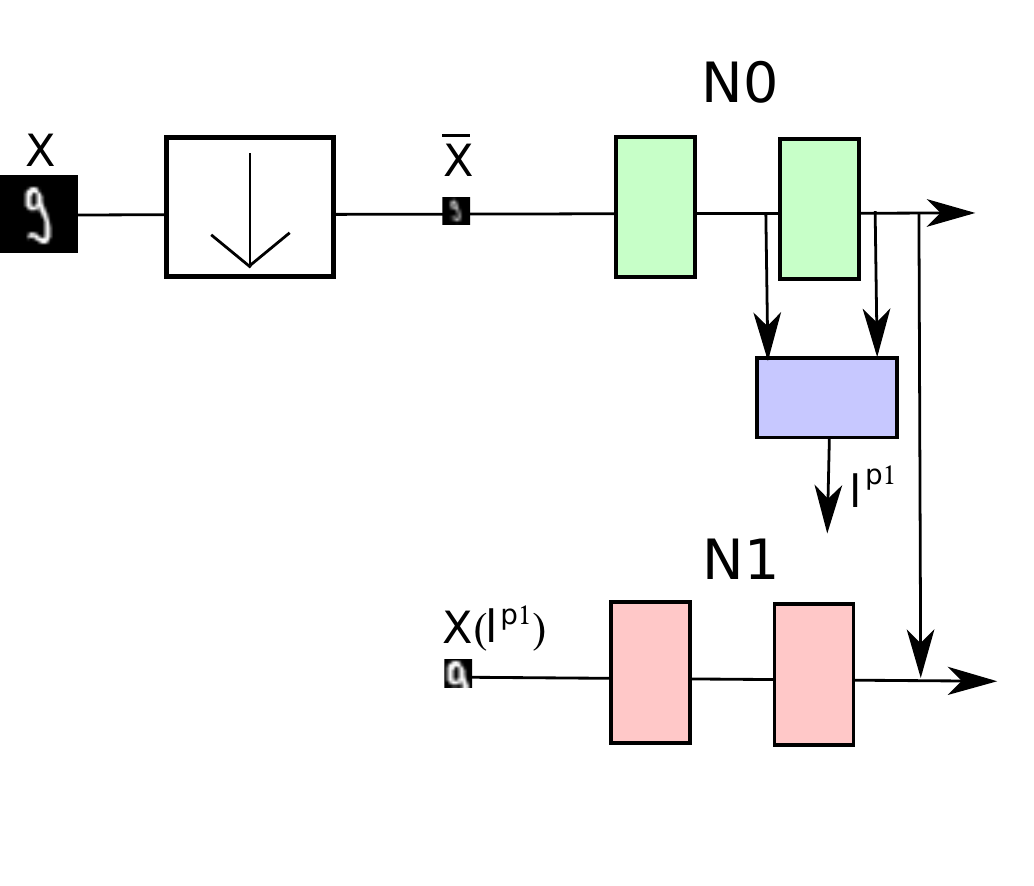}
\end{center}
\caption{Outline of the model at inference time (showing only the
  first glimpse). The green blocks refer to the low resolution
  network, the purple block is used to predict the location where to
  look next and the red blocks represent the first glimpse network
  (taking as input the patch centered at the predicted location). The
  patch shown in the figure was predicted by a model trained on MNIST.}
\label{fig:outline}
\end{figure}

We then feed the patch $X(\l^{p_1})$ to another neural network. Again, for the
sake of simplicity we will refer to a one hidden layer neural
network here. This network, which we call the first ``glimpse
network'' (N1), computes:
\begin{eqnarray}
\begin{array}{lll}
\h^{H_1} & = & \max(0, W_1^{H_1} X(\l^{p_1}) + \b_1^{H_1}) \\
\o^{H_1} & = & W_2^{H_1} \h^{H_1} + \b_2^{H_1}.
\end{array}
\end{eqnarray}
The resulting distribution over the class labels is given by the
geometric mean between the output of the low-resolution network
and the output of the first glimpse network:
\begin{equation}
\y^{H_1} = \mbox{softmax}(\o^{L} + \o^{H_1}) \label{eq:prob-glimpse1}
\end{equation}
A schematic outline of the system is shown in fig.~\ref{fig:outline}.

More generally, at the $n$-th glimpse the predicted class
distribution is defined as:
\begin{equation}
\y^{H_n} = \mbox{softmax}(\o^{L} + 1 / n \sum_{j=1}^n \o^{H_j}).
\end{equation}
where:
\begin{eqnarray}
\begin{array}{lll}
\h^{H_n} & = & \max(0, W_1^{H_n} X(\l^{p_n}) + \b_1^{H_n}) \\
\o^{H_n} & = & W_2^{H_1} \h^{H_n} + \b_2^{H_n}. \label{eq:class-pred-n}
\end{array}
\end{eqnarray}
Preliminary experiments showed that the choice of the weighting of
the terms in the sum is not very important. This weighting yielded
the best results but uniform weighting worked comparably well.
The prediction of the location produced by the $n$-th glimpse
network is:
\begin{equation}
\l^{p_n} = \sigma(W^{p_n}_1 \h^{L} + W^{p_n}_2 (\o^{L} +
\sum_{j=1}^n \o^{H_j}) + \b^{p_n}) \label{eq:loc-pred-n}
\end{equation}
Note that $W_1^{H_n}$, $W_2^{H_n}$, $\b_1^{H_n}$, $\b_2^{H_n}$,
$W^{p_n}_1$, $W^{p_n}_2$ and $\b^{p_n}$ are parameters of the
$n$-th glimpse network (Nn) which are subject to learning at training time.

These equations can be interpreted as follows. The system
aggregates its predictions by taking a geometric mean of the class
predictions at each glimpse (weighting more the prediction of the
low resolution network). Taking a geometric mean amounts to multiplying probability estimates: the overall score is high only when several predictors agree about the presence of the object (akin to an "and" operation).
The location is predicted by taking into
account both low level features and the current prediction of
the (un-normalized) class distribution.
In our preliminary experiments, this configuration yielded the
best results. However, the system is fairly robust to the kind of
average used to integrate information across the glimpses as well as
to the details of how predictions are computed (for instance, on
MNIST the error rate increases by only 0.1\% when using $\y^{H_j}$ instead
of $\o^{H_j}$ for the prediction of where to look next).

\subsection{Learning}
Since we are given a dataset that does not have target values for the locations
where to look, it is going to be challenging to design an algorithm
that can back-propagate through the cropping operator, because
this operator is not differentiable. In this section, we are going to
describe our solution to this problem in an incremental way.

First, we restrict ourselves to the problem of predicting the first
location only and we observe that the location of where to look next can be
interpreted as a latent variable, a variable upon which the loss
depends but that it is not observed.

For each training sample $X$, we define the loss function as a compound sum of terms:
\begin{equation}
\L(\theta_0, \theta_1; X, \z, y^*) = E_0(\bar{X}, y^*; \theta_0) +
E_1(X(\z), y^*;
\theta_1) + \frac{\lambda}{2} || \l^{p_1}
- \z||^2
\end{equation}
where $\theta_0$ collectively denotes the parameters of the low
resolution network $\{ W_1^L, W_2^L, \b_1^L, \b_2^L, W^{p_1}_1, W^{p_1}_2, \b^{p_1} \}$, similarly $\theta_1$
denotes the parameters of the first glimpse network
$\{W_1^H, W_2^H, \b_1^H, \b_2^H \}$, $X(\z)$ is the high resolution
patch extracted at location $\z$,  $\l^{p_1}$ is the predicted patch,
$\bar{X}$ is a low resolution version of the input image $X$, $y^*$ is
the label of the input image and $\lambda$ is a positive scalar.

The first term in the loss is the cross-entropy error of the low
resolution network N0; if the correct class is $k$, then:
\begin{equation}
E_0(\bar{X}, y^k; \theta_0) = - \log(\y^L_k)
\end{equation}
where $\y^L_k$ is the probability assigned to the $k$-th class by the
low resolution network (see eq.~\ref{eq:prob-lowreznet}). Similarly,
the second term in the loss is the cross entropy error of the first
glimpse network (using eq.~\ref{eq:prob-glimpse1} as prediction of the
output class). Finally, the last term measures the discrepancy between
the location $\z$ and the predicted one $\l^{p_1}$.

Similarly to generalized EM and to the Predictive
Sparse Decomposition algorithm~\cite{koray-psd-08}, we train by
alternating between two steps. In the first step (the E step) we minimize the
loss over the latent variables~\footnote{In a probabilistic setting,
this corresponds to computing a MAP estimate of the latent
variables.}:
\begin{equation}
\z^* = \arg \min_{\z} \L(\theta_0, \theta_1; X, \z, y^*) \label{eq:bestz}
\end{equation}
This is the location which increases the log probability of the
correct class while being as close as possible to the current
prediction of the model. Since the cropping operation is not differentiable, we perform
the minimization by a {\em local perturbation}: given the prediction of
the model, we evaluate the loss at nearby image locations. In
practice, we define a grid of points around the initial prediction
(for instance, a 3 by 3 grid centered at the predicted location with
points in the grid stepped every two pixels):
\begin{equation}
\z^* = \arg \min_{\z \in N(\l^{p_1})} \L(\theta_0, \theta_1; X, \z, y^*)
\end{equation}
where $N(\l^{p_1})$ is the discrete set of locations.

In the other step we minimize (or actually do a single
step of gradient descent) over the parameters of the model fixing the
latent variables to the value found at the previous step:
\begin{eqnarray}
\theta_0 \leftarrow  \theta_0 - \eta \frac{\partial \L(\theta_0,
  \theta_1; X, \z^*, y^*) }{\partial \theta_0} \\
\theta_1 \leftarrow  \theta_1 - \eta \frac{\partial \L(\theta_0,
  \theta_1; X, \z^*, y^*) }{\partial \theta_1}
\end{eqnarray}
where $\eta$ is a scalar learning rate. After this gradient step, the
network will produce a better prediction of the location where to
look next and it will increase the log probability of the correct
class.

The minimization over the location variables seems rather wasteful
since only the location yielding the lowest loss is retained. We have
found empirically (see sec.~\ref{sec:expcontrol}) that accuracy
considerably improves by considering also the
``most offending'' location~\cite{EBMtutorial}, that is the location where
the network is most confident about a class that is incorrect:
\begin{equation}
\z^- = \arg \min_{\z \in N(\l^{p_1}), y \neq y^*} \L(\theta_0,
\theta_1; X, \z, y) \label{eq:offendingz}
\end{equation}
We can now add another term to the loss function which is the
cross-entropy at location $z^-$; after the minimization we have:
\begin{equation}
\L(\theta_0, \theta_1; X, y^*) = E_0(\bar{X}, y^*; \theta_0) +
E_1(X(\z^*), y^*; \theta_1) + E_1(X(\z^-), y^*; \theta_1) +
\frac{\lambda}{2} || \l^{p_1} - \z^*||^2 \label{eq:loss-glimpse1}
\end{equation}
This forces the system to increase the probability of the correct
class at the most confusing location as well. This improves generalization
because the location predictor may not be very accurate (overall
if $\lambda$ is small) and this makes the system predict the
correct class not only at the best location but also at nearby ones.

\subsubsection{Extension to Multiple Glimpses}
For the sake of simplicity, we propose to train each glimpse in sequence.
Once, the first glimpse is trained using the algorithm described
above, we train the second glimpse network while holding fix the
parameters of the low resolution network as well as the parameters
of the first glimpse network. More generally, given the first
$k-1$ glimpses we train the $k$-th glimpse network holding fix all
the other parameters. This greedy algorithm is very simple and
worked well in our experiments.

The learning algorithm is the same as the one described above, alternating
between a step which makes a discrete search over locations, and a
step which makes a gradient step over the parameters.

The location is predicted using eq.~\ref{eq:loc-pred-n} while the
output of the glimpse network is given by
eq.~\ref{eq:class-pred-n}. Since we fix the parameters of the
first $k-1$ glimpse networks, the loss reduces to:
\begin{equation}
\L(\theta_k; X, y^*) = E_k(X(\z^*), y^*; \theta_k) + E_k(X(\z^-), y^*; \theta_k) +
\frac{\lambda}{2} || \l^{p_k} - \z^*||^2 \label{eq:loss-glimpsek-a}
\end{equation}

In our preliminary experiments, we have found that on certain
datasets classification
accuracy slightly improves when we constrain the location of each glimpse
to be far away from the locations of previous glimpses. In order
to achieve this, we add to the loss function a penalty for making
predictions that are too close to each other using the following term:
\begin{equation}
 \gamma \sum_{j=1}^{k-1} e^{-\frac{1}{2} \frac{|| \z - \l^{p_j} ||^2}{\sigma^2}} \label{eq:glimpses-term}
\end{equation}
This effectively puts a Gaussian bump penalty around each previously
visited position, encouraging the new prediction to be in a
different position.
 
\subsubsection{Fine Tuning} \label{sec:finetuning}
Since the prediction is never optimal (i.e., yielding the
minimum of the loss function) unless $\lambda$ is very large, in
the last stage of training we set the location to the predicted
value (effectively setting $\lambda$ to infinity) and update only 
the parameters. This tunes the
parameters to fit the actual prediction that will be used at test time.

\subsubsection{Other Extensions}
There are several other extensions to the baseline model:
\begin{itemize}
\item using multi-scale inputs when extracting high-resolution
  patches in order to cheaply leverage larger context (akin to a
  crude approximation of the log-polar resolution of the fovea~\cite{Metamers,Larochelle-fovea}),
\item using gating interactions to modulate the features
  extracted by the glimpse network to take into account also the
  location where the patch was extracted,
\item using gating interactions to predict the location (as
  opposed to doing a simple weighted sum),
\item predicting at every step $M$ locations in parallel as
  opposed to only one,
\item as training proceeds, reducing the grid size used for searching over locations.
\end{itemize}
In sec.~\ref{sec:experiments} we will evaluate some of these extensions.

\subsubsection{Summary}
To summarize, training proceeds by:
\begin{enumerate}
\item train N0 and N1
\item train N2, fixing N0 and N1
\item train N3, fixing N0, N1 and N2
\item etc.
\item fine-tune: using the predicted location, adjust the
  parameters of the last glimpse network to minimize the loss.
\end{enumerate}
The training of each glimpse is done by stochastic gradient
descent with momentum set to 0.9. For each training sample, we
alternate between:
\begin{itemize}
\item finding the location that yields the lowest loss as well as the
  location that yields the most confident but wrong prediction
  (see eq.\ref{eq:bestz} and \ref{eq:offendingz}).
\item updating the parameters by one step of gradient descent using
  the loss in eq.~\ref{eq:loss-glimpsek-a}.
\end{itemize}

\section{Relation to Other Models}
In this section, we relate the proposed method to some prior work.
 First, there has been a lot of
work on methods that produce saliency maps~\cite{Itti98, Kanan10}. Although these
works share the same aim, that is to cut the computational cost of
analyzing very large images by focussing on the most informative regions, they
compute such maps in a task-independent manner and they typically
rely on heuristics~\cite{Lihi-saliency}.

The proposed method is certainly related to the work by Larochelle and
Hinton~\cite{Larochelle-fovea}. Both methods learn all parameters and they are task
driven. Both methods aggregate information across glimpses by
using the geometric mean of the predictions across
glimpses. However, our method is trained in a fully discriminative
manner using a rather different loss and optimization
method. Also,
we do not overlay a grid over all possible locations in
the image but use local perturbations which can be computed more efficiently.  

The particular learning algorithm we chose was inspired by
the Predictive Sparse Decomposition (PSD) algorithm~\cite{koray-psd-08}. Similarly
to this sparse coding algorithm, training alternates between two
steps. In the first step we find the most likely state of some
latent variables (in PSD these are the sparse codes while in this work
these are variables encoding the spatial position of where to look
next at high resolution). Notice however that the optimization
technique we employ is drastically different: in this work we
minimize by a local discrete perturbation because the objective is
not differentiable, while in that work they minimize by
gradient descent. Also, in this work the loss includes a
contrastive term which is absent in PSD. In the second step, the parameters are
updated by gradient descent in both algorithms.
Although the design of the loss function may resemble the one
proposed here, the task and the meaning of the variables is
very different: in~\cite{koray-psd-08} the objective is reconstructing the
input while here the objective is to classify, in~\cite{koray-psd-08}
the variables represent internal features while here they
represent spatial locations.

If the glimpse networks share the same parameters (except for the
module that predicts the location) and we augment the model with
gating units that make the hidden units depend on the selected
location, then the proposed method is also related to Recurrent
Neural Networks (RNN). The major difference is that in this model the input
which is fed to the system is self-determined at each step, as opposed to be
provided by the user.

Finally, the proposed method can also be related to cascade
approaches and boosting~\cite{viola-01}. As we shall see in the next section, it
is possible to set a threshold on the confidence of the
classifiers and use that to make an amount of computation which
depends on the complexity of the input. In fact, most input images
exhibit patterns that are simple enough to be classified well by
the low resolution network itself. Fewer and more complex patterns instead require the
processing of the subsequent glimpse networks. This approach which
is reminiscent of cascading, makes classification very fast and
adaptive to the complexity of the input, as we shall see in sec.~\ref{sec:exp-jmnist}.

\section{Experiments} \label{sec:experiments}
\begin{figure}[!t]
\begin{center}
\includegraphics[width=.8\linewidth]{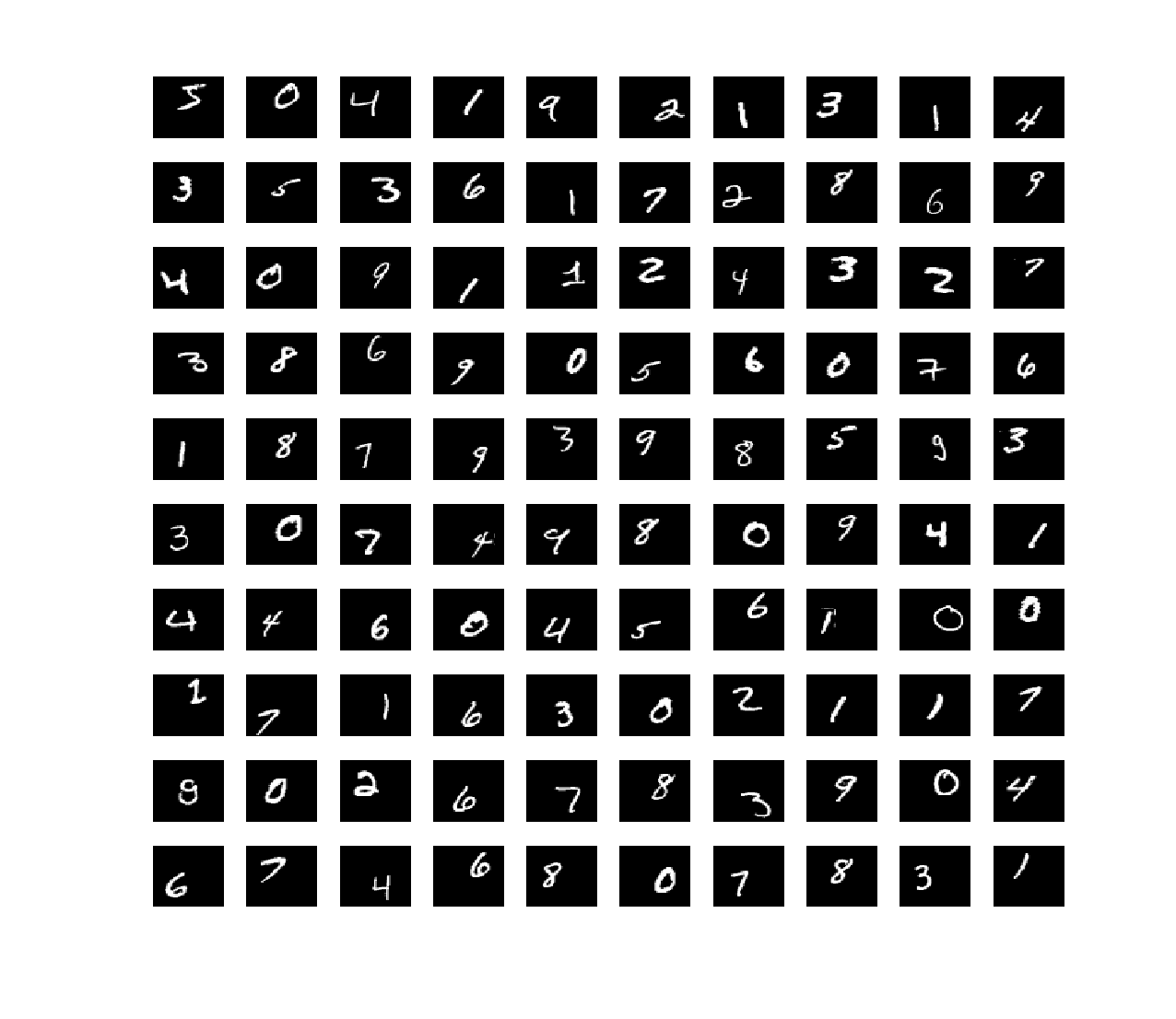}
\end{center}
\caption{Examples of images in the jittered MNIST dataset. Each image
  has size 48 by 48 pixels and digits are placed at random locations
  in the image.}
\label{fig:mnist_jittered.png}
\end{figure}
Most of our experiments are reported on a variant of the MNIST
dataset~\cite{MNIST}: the jittered MNIST dataset. This dataset consists of 48
by 48 pixel images that were constructed by placing at a random
location the original 28 by 28 pixel image and by filling with zeros the
remaining pixel values. A random subset of the images in this
dataset is shown in fig.~\ref{fig:mnist_jittered.png}.
The training set consists of 600,000 images (each image from the
original MNIST training set generates 10 images with the same digit at different random locations) and 30,000 images in the test
set.

The advantage of this dataset is that the system is less prone to
overfit and therefore, we can better assess the quality of
different training strategies without worrying too much about the
confounding factor of regularization.

For the sake of comparison, we also report experiments on the
original MNIST dataset. These experiments demonstrate that the
proposed method perform as well or very close to convolutional
networks (which are the state-of-the-art classification method on
this dataset), while being several times more
computationally efficient. 

\subsection{Jittered MNIST} \label{sec:exp-jmnist}
\begin{table}[!b]
\caption{Results on the jittered MNIST dataset.
 A speed-up equal to 12 means that the
  corresponding method makes 12 times less flops than the baseline
  fully 
  connected network operating on the original high resolution images.
}
\begin{center}
\begin{tabular}{l|c|c}
{\bf Method} & {\bf Test Error Rate\%} & {\bf Speed-Up} \\
\hline
\hline
fully connected on high resolution images & 3.5 & 1.0 \\
\hline
convolutional network & 1.4 & 0.03 \\
\hline
this method, 1 glimpse & 2.3 & 5 \\
\hline
this method, 2 glimpses & 1.6 & 3 \\
\hline
this method, 2 glimpses, fine tuned & 1.4 & 3 \\
\hline
this method, 2 glimpses, cascaded & 1.7 & 12 \\
\end{tabular}
\end{center}
\label{tab:jmnist-results}
\end{table}
In this experiment, the input images have size 48 by 48 pixels.
These are down-sampled to 12 by 12 pixels before being fed to the
low resolution network N0. The high resolution patches have also size
12 by 12 pixels and we consider two scales: the original one and
half of the original one. In other words, the glimpse networks
take as input two patches of size 12 by 12 pixels, where the
second patch is actually a down-sampled version of a 24 by 24
patch.

During training, the search over spatial locations is performed
over a 3 by 3 grid of points that are spaced every 2 pixels (as
measured at the finest resolution). We tried other patch sizes,
grid sizes, etc. and we found that the performance is fairly
robust to the choice of these hyper-parameters.
In all our experiments we set $\lambda$ to 100, $\gamma$ to 0.01
and $\sigma^2$ to 0.002 (see eq.~\ref{eq:loss-glimpsek-a} and
\ref{eq:glimpses-term}). 
Finally, during training we run for 50 sweeps over the training set using
stochastic gradient descent on mini-batches of size 50 with
momentum set to 0.9 and learning rate set to 0.05.

Tab.~\ref{tab:jmnist-results} reports the results in terms of
error rate and speed up with respect to the number of flops
required by a fully connected network trained on the original 48
by 48 images. All networks we consider (N0, N1, N2 as well as the
fully connected baseline) have one hidden layer with 500 hidden units.
We also compare to a convolutional network which is the best
classification system on this data. The convolutional network has
two stages of convolution and max-pooling, each of which with 64
filters of size 5 by 5 and with pooling neighborhood of 2 by 2
pixels stepped every 2 pixels. These two stages
are followed by a fully connected layer with 128 hidden units and
by a top layer with 10 units, as many as the classes we want to predict.

We found that adding glimpses helps performance (a third glimpse
yielded only marginal improvements though). Moreover, fine-tuning
(as described in sec.~\ref{sec:finetuning}) further reduces the
error rate and makes the system match the performance of the
convolutional network. Notice however that the amount of
computation required by our method is 2 order of magnitude lower
than what is required by a convolutional network.

The last row of tab.~\ref{tab:jmnist-results} shows the
performance of our system when using a threshold on the confidence
of the prediction in order to further cut the computational cost.
By setting a threshold of 0.95 on the output of the classifiers,
we have that 78\% of the test samples can be classified by just the low
resolution network. Among the classified samples, we have
have an error rate of 0.45\% and an overall rejection rate of
22\%. The first glimpse network classifies 15\% of the digits (relative to the
total number of test samples) and the second glimpse network 3\%
of the digits with a final rejection rate of 4.8\%.
If we let N2 classify all digits regardless of its confidence, the
overall test error rate becomes 1.7\%, only 0.1\% worse 
than the reference model (which runs all samples through all the
networks N0, N1 and N2). Since most
samples are confidently classified by N0, the computational
saving is very large: 12 times more efficient than using the fully
connected baseline.

\subsubsection{Control Experiments} \label{sec:expcontrol}
\begin{table}[!b]
\caption{Control experiments on the jittered MNIST dataset.}
\begin{center}
\begin{tabular}{l|c|c}
{\bf Method} & {\bf Test Error Rate\%} & {\bf Speed-Up} \\
\hline
\hline
this method, 1 glimpse (multi-resolution) & 2.3 & 5 \\
\hline
this method, 1 glimpse (single resolution) & 4.2 & 7 \\
\hline
this method, 1 glimpse (single resolution), no contrastive term & 14.0 & 7 \\
\hline
\hline
this method, 2 glimpses & 1.6 & 3 \\
\hline
this method, 2 glimpses, done in parallel & 1.75 & 3 \\
\hline
this method, 2 glimpses, no diversity penalty & 1.4 & 3 \\
\hline
this method, 2 glimpses, no diversity penalty with fine tuning & 1.2 & 3 \\
\hline
this method, 2 glimpses, with weight sharing & 1.7 & 3
\end{tabular}
\end{center}
\label{tab:jmnist-results-control}
\end{table}
In this section, we further investigate the proposed model 
teasing apart the contribution of each component of the system.
The experimental set up is the same as the one described in the
previous section. In the first experiment, we validated the
effectiveness of the use of multi-scale inputs in N1. As
the first two rows of tab.~\ref{tab:jmnist-results-control} show,
the test error rate is 
greatly reduced by providing the network with more context. We
conjecture that this makes the network also more robust to errors
in the exact prediction of the location where to look next.

In the next experiment we remove the contrastive term in the loss
function which measures the cross entropy at the most offending
location (see third term on the right hand side of
eq.~\ref{eq:loss-glimpse1}). This dramatically increases the error
rate, demonstrating that it is very important to lower the
score of incorrect labels at nearby locations.

In the fourth and fifth row of tab.~\ref{tab:jmnist-results-control} we
report the error rate when the two glimpses are performed
sequentially and in parallel. In the latter experiment, the prediction
of the location of the first and second glimpses takes the same
input, namely the first hidden layer and the output of N0.
This experiment aims at assessing the importance of the sequential
nature of the inference process we proposed. The error rate does
increase but not very much, suggesting that the greedy algorithm we
proposed is better but we would not compromise too much accuracy
by performing predictions in parallel. This suggests that we may
reach a better trade-off between computational efficiency and
accuracy by performing $M$ sequential predictions, each composed
by $K$ parallel glimpses. We leave further exploration of this
conjecture to future investigation. 

In the next experiment, we verify the effectiveness of the error
term penalizing predictions of nearby locations (see
eq.~\ref{eq:glimpses-term}). By removing this term the error rate
actually decreases by 0.2\%, suggesting that this term could be
safely removed from the loss~\footnote{However, on the original MNIST dataset
  the error rate increases by 0.1\% when removing this term.}.
If we fine tune the last glimpse network using this set up, we
reach an error rate of 1.2\%, which is the lowest error rate we
could report on this dataset.

Finally, we report the error rate when the parameters in N1 and N2
are tied to each other. The last row of
tab.~\ref{tab:jmnist-results-control} shows that using the same
parameters for different glimpses increases the error rate by only
0.1\%.

\subsubsection{Visualizations}
\begin{figure}[!t]
\begin{center}
\includegraphics[width=.8\linewidth]{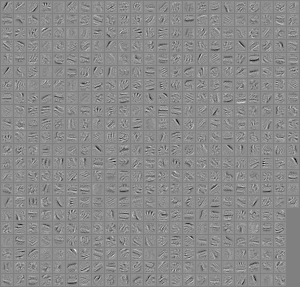}
\end{center}
\caption{Each tile is a filter learned by the low resolution network N0.}
\label{fig:many_multirez_f2_sz12_l9_s2_lr01_wpos100_low_rez_filters.png}
\end{figure}

\begin{figure}[!t]
\begin{center}
\includegraphics[width=.8\linewidth]{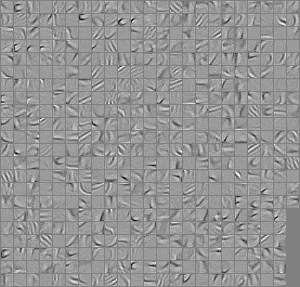}
\end{center}
\caption{Each tile is a filter learned by the first glimpse network
  N1 (applied to high resolution patches).}
\label{fig:many_multirez_f2_sz12_l9_s2_lr01_wpos100_high_rez_filters.png}
\end{figure}
In this section, we provide qualitative intuitions as to what the
system is
doing. Fig.~\ref{fig:many_multirez_f2_sz12_l9_s2_lr01_wpos100_low_rez_filters.png}
and
\ref{fig:many_multirez_f2_sz12_l9_s2_lr01_wpos100_high_rez_filters.png}
show the filters (i.e., weights connecting each hidden unit to the
corresponding input) learned by N0 and N1. All filters seem to be
used and they converge to highly tuned stroke detectors. We
conjecture that the network performs its classification by doing a
global pattern matching using the low resolution images, and 
a highly tuned local feature matching using N1 and N2.

\begin{figure}[!t]
\begin{center}
\includegraphics[width=.5\linewidth]{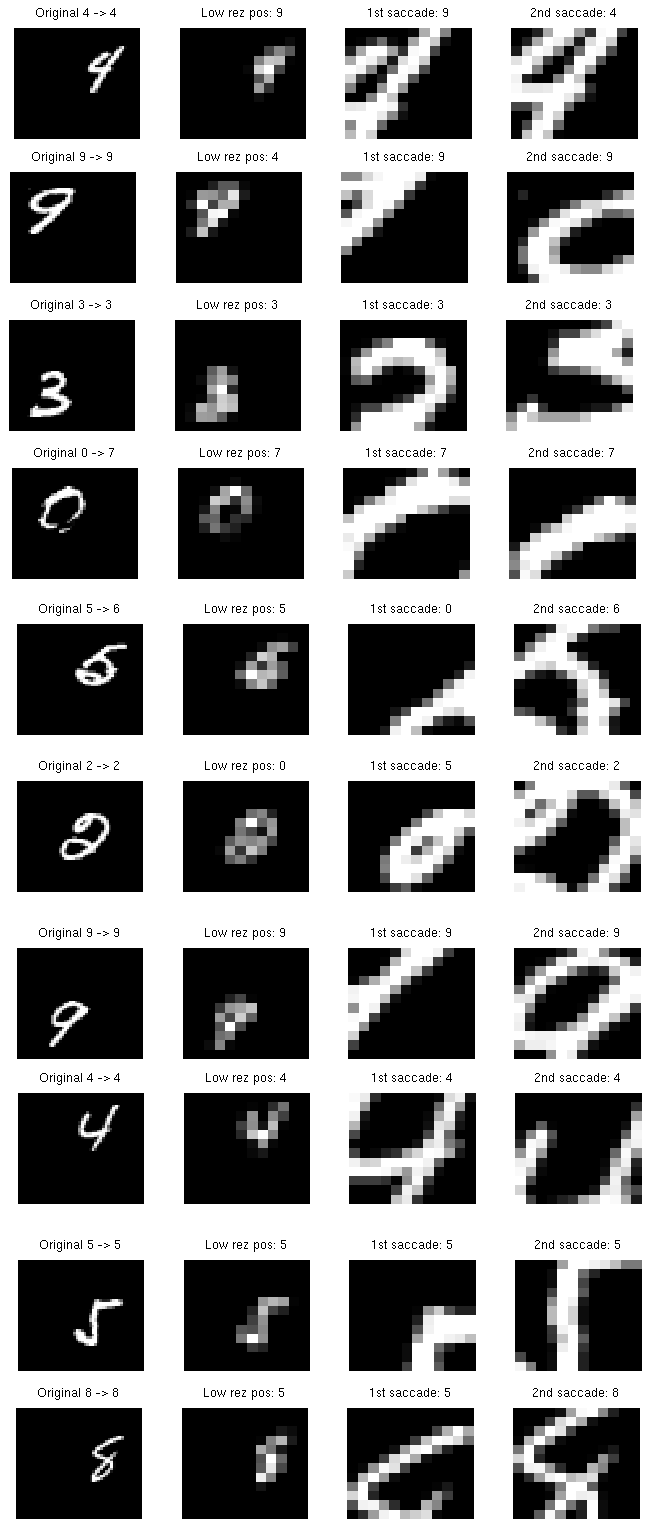}
\end{center}
\caption{Examples of predictions of test images. Every row shows a
  different example. The columns show (from left to right): the
  original image $X$ (with the correct class label and the predicted class
  label), the down-sampled image $\bar{X}$ of size 12 by 12 pixels
  (with the predicted class produced by N0 network), the high
  resolution patch used by the first glimpse network N1 (with the
  its class prediction) and the high resolution patch fed to the
  second glimpse network N2 (with the corresponding class label
  prediction). Note that each image has been resize to fit the grid,
  but the images in the first column have size 48 by 48, while the
  images in the other columns have size 12 by 12 pixels.}
\label{fig:example_predictions_jmnist}
\end{figure}
In fig.~\ref{fig:example_predictions_jmnist} we show examples were
the predictions produced by N0, N1 and N2 differ the most.
First, we observe that N1 and N2 look at different locations in order to
disambiguate the low resolution image (for instance, N1 looks at
the top part of 3 and 8 while N2 looks at the bottom part of these
images). Second, the prediction is often refined as glimpses are added
and evidence aggregated. And finally, the predicted location
tracks well the digit as it moves in the canvas.

\subsection{MNIST}
\begin{table}[!b]
\caption{Results on MNIST dataset; the speed-up is relative to the fully
  connected network.  Original images have size 28 by 28 pixels,
  down-sampled images have size 10 by 10 pixels, high resolution
  patches have size 10 by 10 pixels. Notice that the reported
  results of~\cite{Larochelle-fovea} was achieved by training on 10,000 training
  samples only. Our method achieves the same accuracy when trained
  on such a small subset of the training data. According to
  Larochelle, their method~\cite{Larochelle-fovea} does not scale easily to the full
  size of the dataset.}
\begin{center}
\begin{tabular}{l|c|c}
{\bf Method} & {\bf Test Error Rate\%} & {\bf Speed-Up} \\
\hline
\hline
fully connected on high resolution images & 1.8 & 1.0 \\
\hline
fully connected on down-sampled images & 7.7 & 7.2 \\
\hline
this method, 1 glimpse fixed at the center of the image & 2.0 & 3.6 \\
\hline
convolutional network & 0.8 & 0.05 \\
\hline
prior work~\cite{Larochelle-fovea} & 2.7 & N/A \\
\hline
this method, 1 glimpse & 1.2 & 3.6
\end{tabular}
\end{center}
\label{tab:mnist-results}
\end{table}
In the last experiment, we report experiments on the original
version of the MNIST dataset in order to compare to
previously published results~\cite{MNIST}. In this experiment, the input images
have size 28 by 28 pixels while the down-sampled images have size
10 by 10 pixels. The size of the patches that we crop from the
high resolution images is also 10 by 10. Given the small size of
the original images in this dataset, we report results using only a single resolution
(the finest). During training the search over locations is
performed by considering a 3 by 3 grid of points strided every 2
pixels. 
Finally, during training we run for 50 sweeps over the whole training set using
stochastic gradient descent on mini-batches of size 50 with
momentum set to 0.9 and learning rate set to 0.01.  

Tab.~\ref{tab:mnist-results} compares our method to standard
baselines (fully connected network taking the original and
the down-sampled input as well as a convolutional network). We also
compare to our method with a fixed glimpse at the center of
the image (since the MNIST digits are registered, the central
patch is the most informative one). We observe that our
method generalizes better than a fully connected network, and it
approaches the performance of a convolutional network (without
adding any regularization) when we learn where to look in the
image. A second glimpse produced only marginal gains (a reduction
of 0.1\% test error rate) due to severe overfitting. 

\section{Conclusions}
In this work, we introduced a new model to classify images
through a sequence of glimpses. The whole system is trained in
a discriminative manner by alternating between two steps. 
In the first step, we find the best locations where to look next
as measured by the increase in the likelihood of the image
belonging to the correct class. In the second step, the parameters
of the model are updated in order to improve both prediction of
where to look next as well as prediction of the object category.

After training, the model is computationally very efficient since
a simple forward pass yields the class prediction. Moreover, the
sequential processing of the method can be leveraged to further
cut its computational cost: the next glimpse is used only when the
current prediction has low confidence. We have shown that such a
method perform comparably well to state-of-the-art classification
methods at a tiny fraction of their computational cost.

Future work will investigate the use of this method on more
challenging benchmarks for both classification and detection.

\section{Acknowledgements}
I would like to thank Geoff Hinton for encouraging me to work on
this project and for his advises. I would like
to thank also Ilya Sutskever and Jay Yagnik for very helpful
discussions and suggestions, and Andrew
Senior and Alex Krizhevsky for their help in running experiments on GPU.
Finally, this project could not have
been accomplished without the support from Jeff Dean and the
Google Brain team at Google.
\bibliographystyle{unsrt}

\bibliography{ranzato_fixationsApr2014}

\end{document}